%% file: main.tex
\documentclass[runningheads]{llncs}

\usepackage{eccv}

\usepackage{eccvabbrv}
\usepackage{multirow}

\usepackage{graphicx}
\usepackage{booktabs}
\usepackage{comment}
\usepackage{todonotes}
\usepackage{algorithm}
\usepackage{algpseudocode}
\usepackage{diagbox}
\usepackage{graphicx}
\usepackage{amsmath}
\usepackage{amssymb}
\usepackage{booktabs}
\usepackage{algorithm}
\usepackage{algpseudocode}
\usepackage{listings}
\usepackage{xcolor}
\usepackage{wrapfig} 

\usepackage{pifont}
\usepackage[accsupp]{axessibility}  

\usepackage{xcolor,colortbl}
\definecolor{Greener}{RGB}{210, 253, 187}
\definecolor{LightGreen}{RGB}{236, 255, 239}

\newcommand{\paragraphOur}[1]{\vspace{2mm}\noindent\textbf{#1}}

\usepackage{wrapfig}
\newcommand{\greenbox}[1]{\fboxsep=1pt\fboxrule=2pt\fcolorbox{green}{white}{#1}}
\newcommand{\redbox}[1]{\fboxsep=1pt\fboxrule=2pt\fcolorbox{red}{white}{#1}}

\usepackage{hyperref}

\usepackage{orcidlink}

\begin{document}

\title{WildFusion: Individual Animal Identification \\ with Calibrated Similarity Fusion} 


\author{First Author\inst{1}\orcidlink{0000-1111-2222-3333} \and
Second Author\inst{2,3}\orcidlink{1111-2222-3333-4444} \and
Third Author\inst{3}\orcidlink{2222--3333-4444-5555}}

\authorrunning{F.~Author et al.}

\titlerunning{Individual Animal Identification
with Calibrated Similarity Fusion}

\author{
Vojtěch Cermak\inst{1}\orcidlink{} \and
Lukas Picek\inst{2,3}\orcidlink{0000-0002-6041-9722} \and
Luk\'a\v{s} Adam\inst{4}\orcidlink{0000-0001-8748-4308} \and
\\ Lukáš Neumann\inst{1}\orcidlink{0000-0002-9428-3712} \and
Jiří Matas\inst{1}\orcidlink{0000-0003-0863-4844}}

\authorrunning{V. Cermak et al.}

\institute{
Czech Technical University in Prague, FEL, CMP, Prague, Czechia\\ \and
University of West Bohemia, FAS, Department of Cybernetics, Pilsen, Czechia \and 
Inria, LIRMM, University of Montpellier, CNRS, Montpellier, France\\ \and 
University of West Bohemia, FEE, RICE, Pilsen, Czechia \\
\email{\{cermavo3,matas\}@fel.cvut.cz, lukaspicek@gmail.com, adamluk3@fel.zcu.cz}}
\maketitle

\begin{abstract}
We propose a new method -- WildFusion -- for individual identification of a broad range of animal species.
The method fuses deep scores (e.g., MegaDescriptor or DINOv2) and
local matching similarity (e.g., LoFTR and LightGlue) to identify individual animals.
The global and local information fusion is facilitated by similarity score calibration.
In a zero-shot setting, relying on local similarity score only,  WildFusion achieved mean accuracy,
measured on 17 datasets, of 76.2\%.
This is better than the state-of-the-art model, MegaDescriptor-L, whose training set included 15 of the 17 datasets.
If a dataset-specific calibration is applied, mean accuracy increases by 2.3\% percentage points.
WildFusion, with both local and global similarity scores, outperforms the state-of-the-art significantly --
mean accuracy reached 84.0\%, an increase of 8.5 percentage points;  the mean relative error drops 
by~35\%.
We make the code and pre-trained models publicly available\footnote{\url{https://github.com/WildlifeDatasets/wildlife-tools}}, enabling immediate use in ecology and conservation.
\end{abstract}

\section{Introduction}
\label{sec:intro}

Identifying individual animals is essential in various domains of wildlife research. It help us understand the complexities of species dynamics\cite{vidal2021perspectives, papafitsoros_2021}, which is necessary for developing efficient conservation strategies. Besides, it can improve the accuracy of population density estimation, which is important in problems like disease monitoring and control\cite{palencia2023not}, the role of the animal in the ecosystem\cite{rowcliffe2008estimating}, monitoring invasive species\cite{caravaggi2016invasive} and measuring the involvement of humans in the animal's habitat and ecological restoration\cite{blount2021covid}.
Accurate identification requires domain knowledge and is extremely time-consuming due to the need for manual data processing.
Therefore, considerable progress has been made in the development of methods for automating this process. Even though identifying animal individuals from images is challenging, machine learning and computer vision methods applied to species with unique patterns already enhance ecological research\cite{weissbrod2013automated}.
The automation of animal re-identification is typically based on (i) \textit{deep learning}, (ii) \textit{local feature matching}, or (iii) \textit{species-specific methods}. 

\begin{figure}[t]
    \centering
    \includegraphics[width=0.975\linewidth]{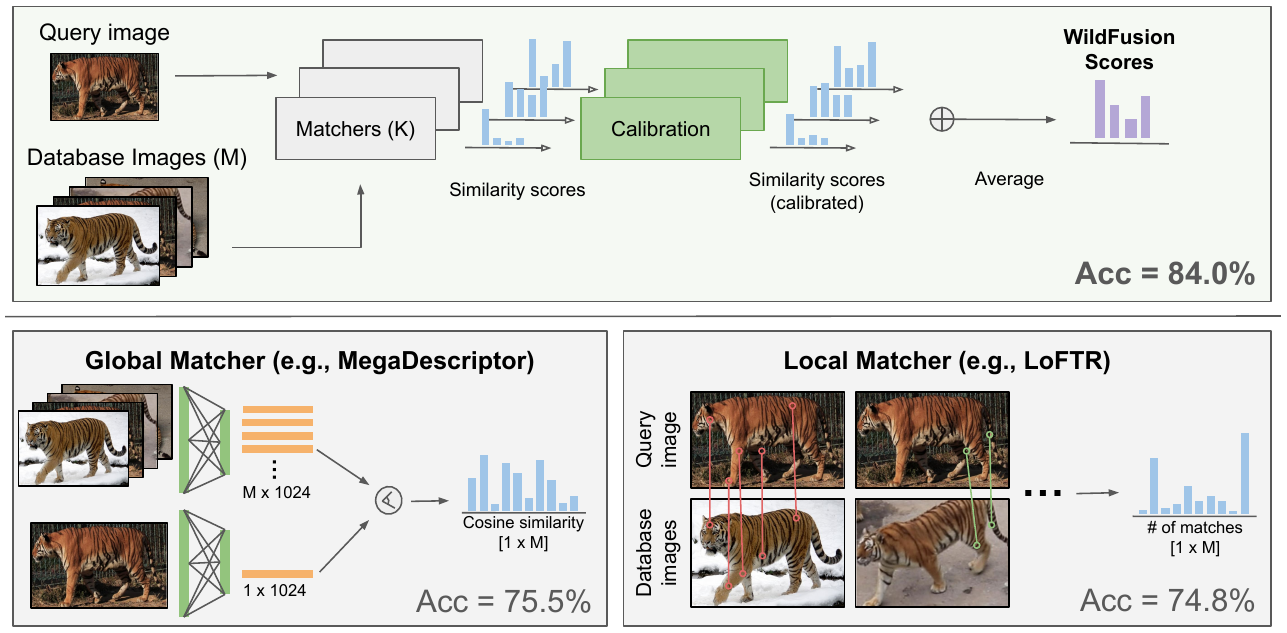}
    \caption{\textbf{Calibrated similarity fusion.} Fusing local (in the $[0, \mathcal{R}]$ range) and global matching scores (e.g., cosine similarity) is not possible without calibration. By calibrating the outputs of any local and global matcher, we can easily fuse them and achieve better performance. In terms of accuracy and evaluated on 17 datasets, we increased the performance by 8.5\% on average and reduced relative error by 35\%.}
    \label{fig:enter-label}
\end{figure}

The \textit{deep learning}-based approaches\cite{ferreira2020deep, bruslund2020re,ueno2022automatic,li2019atrw,deb2018face,miele2021revisiting} use either a standard classification-like approach or metric learning. Even though those approaches perform well, the models need relatively large annotated datasets, and fine-tuning requires considerable computational resources.
On the other hand, methods based on \textit{local descriptors} (e.g., SIFT\cite{lowe2004distinctive}, SuperPoint\cite{detone2018superpoint}) can be used without fine-tuning.\cite{parham2017animal, reno2019sift,andrew2016automatic,dunbar2021hotspotter}. Indeed, the overall accuracy of matching local descriptors does not achieve a performance of deep learning methods\cite{Cermak_2024_WACV}, but those approaches are still very popular due to the existence of open-source tools (e.g., HotSpoter, WildMe) that are based on local descriptors.
Additionally, matching requires a pairwise comparison between all query and database samples. As the identity database grows in size, the computational time quickly becomes unfeasible; therefore, local feature matching remains a viable option only for moderately sized datasets.
The \textit{species-specific} methods are usually tailored to suit species without any visual characteristics\cite{bedetti2020system,weideman2020extracting,gilman2016computer,kelly2001computer,anderson2010computer}. Existing methods focus, for example, on the shape of an elephant's ear, the facial characteristics of primates, or the fluke shape of whales. However, due to their idiosyncratic nature, these methods are difficult to transfer to other species.

In light of that, we propose WildFusion, a new state-of-the-art approach to zero-shot animal re-identification. It fuses calibrated deep similarity functions (i.e., MegaDescriptor and/or DINOv2 feature similarity) and local matching 
scores (i.e., number of matches from descriptors such as LoFTR and LightGlue) to select an identity from a database. For reference, see the illustration in Figure~\ref{fig:enter-label}.
With this straightforward approach, WildFusion significantly outperforms the current state-of-the-art without domain adaptation or fine-tuning. 

\newpage
\vspace{2pt}
\noindent\textbf{The main contributions of this paper are:}
\begin{itemize}
\vspace{-5pt}
    \item A new ensembling framework (WildFusion) that allows a combination of deep- and local-feature matching scores.
    \item A state-of-the-art performance on a set of animal identification problems, outperforming current methods by 8.5\% on average; measured on 17 datasets.
    \item Comprehensive evaluation of selected state-of-the-art deep-learning and local feature-matching methods for image-matching and animal re-identification.
    \item Showing that WildFusion works without the need for fine-tuning and provides state-of-the-art performance out of the box, even in a zero-shot setting.
\end{itemize}

\section{Related work}
\label{sec:related work}
\vspace{-0.25cm}
\paragraphOur{Local feature matching methods}: Early work on animal re-identification used hand-crafted features such as cheetah's spots\cite{kelly2001computer} or zebra's stripes\cite{lahiri2011biometric}. Since these approaches suffer from poor performance and are non-transferable to other species, methods extracting local patterns such as SIFT\cite{lowe2004distinctive} or ORB\cite{rublee2011orb} were soon widely used.
They extract descriptors from a database of images and match the descriptors from image pairs. Popular SW, e.g., WildID\cite{bolger2012computer}, HotSpotter\cite{crall2013hotspotter} and \href{https://github.com/daniel-brenot/I3S-Interactive-Individual-Identification-System-Desktop}{I$^3$S} are using such approach for years. 
Recently, the focus moved to local features extracted by deep networks such as ALIKED\cite{Zhao2023ALIKED}, DISK\cite{tyszkiewicz2020disk} or
SuperPoint\cite{detone2018superpoint}. The classical matching of local descriptors could be simply replaced by deep methods such as LightGlue\cite{lindenberger2023lightglue}, Superglue\cite{sarlin2020superglue}, and LoFTR\cite{sun2021loftr} that allow both extracting and matching of the local features. These matching methods return potential matches and their confidence scores. They require manual thresholding to determine which features are matched. In animal re-identification, deep local features are slowly coming into focus; for example,\cite{pedersen2022re} used a combination of the SuperPoint features with the SuperGlue matching.

\paragraphOur{Deep embedding methods}: The applications of deep methods in animal re-identification are relatively new\cite{carter2014automated}. The simplest use case consists of extracting embeddings from a neural network and feeding them to an SVM classifier\cite{moreira2017my,clapham2020automated,korschens2019elpephants}. This approach has low computational demands, but the network cannot be fine-tuned. Another simple approach involves fine-tuning a pre-trained neural network\cite{ferreira2020deep,schneider2018can}. These approaches usually require a fixed number of classes (individuals), which is not realistic. For this reason, metric learning methods (e.g., ArcFace\cite{Cermak_2024_WACV}, Siamese networks\cite{kabuga2019using}, and Triplet loss\cite{dlamini2020automated,miele2021revisiting}) became popular. Instead of classifying images into a pre-determined set of classes, they measure differences between images and are, therefore, able to generalize into new individuals. Another approach is to use publicly available large-scale, foundational models pre-trained on large datasets such as BioCLIP\cite{stevens2024bioclip}, DINOv2\cite{oquab2023dinov2}, and MegaDescriptor\cite{Cermak_2024_WACV}. Since these models are primarily designed for general computer vision tasks, they are not adapted for the nuances of wildlife re-identification, which heavily relies on fine-grained patterns. 
This was addressed by MegaDescriptor\cite{Cermak_2024_WACV}, the Swin-based foundational model for animal re-identification that was trained on over 30 datasets (collected using \href{https://github.com/WildlifeDatasets/wildlife-datasets}{WildlifeDatasets}) using ArcFace loss\cite{deng2019arcface}.

\paragraphOur{Species-specific methods} are usually tailored to a particular species or closely related species and involve pre-processing steps such as extracting patches from regions of interest or accurately aligning images. Besides, they are not transferable to other species. Examples of these methods are Amphident\cite{drechsler2015genetic} which 
find matching pixels within newt patterns or CurvRank~v2\cite{moskvyak2021robust} and finFindR\cite{weideman2017integral}, which match the fin curvatures to identify mantas, dolphins, or whales.

\section{Methodology}
In this section, we describe the similarity scores based on deep embeddings and local feature matching and introduce the proposed WildFusion, a method for finding an image from a database of images $x_1,\dots,x_D$ closest to query image~$x_q$. \\

\noindent\textit{\textbf{Note}: For wildlife re-identification, we employ a standard setting inspired by practical applications in animal ecology, widely used in automated animal re-id studies\cite{Cermak_2024_WACV,adam2024seaturtleid}.  
This setting corresponds to the image retrieval problem, where the goal is to find the most visually similar images (whose identity is used as prediction) in the database for a given query image based on a similarity metric.}

\subsection{Global similarity score}
Given an image $x$, we use a neural network $f(x)$
to extract a fixed-length embedding. The network $f(x)$ is a complex function that maps images into embedding space where the representations of images depicting the same animal are closer together, while those of different individual animals are distinctively separated. 
Common architectures of neural networks include convolutional\cite{liu2022convnet, next} or transformer-based\cite{dosovitskiy2020image, liu2021swin} architectures and are often trained with metric learning, e.g.,  ArcFace\cite{deng2019arcface} and Triplet loss\cite{schroff2015facenet},
to promote separability in the embedding space. The similarity between images is calculated as the similarity between their representation in the embedding space. Formally, we define the \textit{global similarity} between two images $x_0$ and $x_1$ as the cosine similarity between their corresponding deep embeddings extracted by neural network $f$:
\begin{equation}\label{eq:similarity_deep}
s_G(x_0, x_1) = \frac{f(x_0) \cdot f(x_1)}{\|f(x_0)\| \|f(x_1)|}.   
\end{equation}

\subsection{Matching based similarity score}
We derive a similarity metric based on local feature matching as the number of found significant matches. The feature matching methods return a list of potential matches and their confidence score. We declare a match to be significant if its confidence score is above some threshold $\mu$. Formally, given two images $x_0$ and $x_1$ with the number of matches $M(x_0, x_1)$ with confidence scores $c_m(x_0, x_1)$, we define the \textit{local similarity} metric as
\begin{equation}\label{eq:similarity_local}
s_{L}(x_0, x_1) = \sum_{m=1}^{M(x_0, x_1)} I\big(c_m(x_0, x_1) > \mu\big),
\end{equation}

where $I$ is the counting (0/1) function. This approach requires the tuning of the thresholding hyperparameter $\mu$, where large values of $\mu$ allow for a small number of high-quality matches, while low values of $\mu$ result in a larger amount of matches with potentially lower quality. As we empirically show in Section \ref{sec:effect-mu}, all considered feature matching methods are robust to $\mu$ selections with $\mu=0.5$ being reasonable choice in most scenarios.

\subsection{Score calibration}
Calibration refers to rescaling model outputs so that they could be interpreted probabilistically. In our case, it is required to normalize the outputs of multiple models to the common range $[0,1]$.
Predictions of well-calibrated model reflect confidence in the given class predictions\cite{gao2021towards}. 

We apply calibration to the predicted similarity scores. The similarity scores are used for comparison and ranking in image retrieval, with the magnitude of the scores having no direct interpretation. We use calibration to ensure that the predicted similarity score corresponds to the probability that the images in a pair have the same identity. We construct calibrated scores from either global or matching-based scores using a calibration function $f_{\text{cal}}: \mathbb{R} \rightarrow [0, 1]$.

\begin{equation}\label{eq:calibration}
\hat{s}(x_0, x_1) = f_{cal}(s(x_0, x_1)).   
\end{equation}

A common approach for constructing $f_{\text{cal}}$ is Platt scaling\cite{platt1999probabilistic}, which involves fitting a single-variable logistic regression with uncalibrated scores as inputs. Another widely used method is isotonic regression\cite{zadrozny2002transforming}, a variant of binning regression with a monotonicity constraint. Given uncalibrated scores, it learns a non-decreasing piecewise constant function. However, for our application in ranking and image retrieval, we require a strictly increasing function to handle ties in scores. To achieve this, we first apply the isotonic regression and second interpolate the bin centers using a Piecewise Cubic Hermite Interpolating Polynomial  (PCHIP)\cite{fritsch1980monotone}, which performs cubic interpolation while preserving monotonicity. This procedure results in the required strictly increasing calibration function.

\subsection{WildFusion -- Calibrated score ensembling}
To construct an ensemble, we consider $K$ models with similarity scores $s_i(x_0,x_1)$ for a pair of images $(x_0,x_1)$.  The calibrated scores $\hat s_i(x_0,x_1)$ are interpreted as estimates of same true probability $P(\text{id}(x_0) = \text{id}(x_1) \mid x_0, x_1)$. To denoise the prediction, we assume the probabilities to be independent observations with additive noise. The WildFusion score is thus  a weighted average of $n$ calibrated scores:
\begin{equation}\label{eq}
s_{F}(x_0, x_1) = \sum_{i=1}^{K} w_{i} \hat{s}_{i}(x_0, x_1),
\end{equation}

where the weights should reflect the variance of the additive noise in the score. If we assume that all scores have similar variance, equal weights $w_i=\frac1n$ are selected, and the weighted average reduces to the simple average.


\begin{figure}[!h]
\centering
\begin{tabular}{ccccc}

    \includegraphics[height=1.9cm,width=2.3cm]{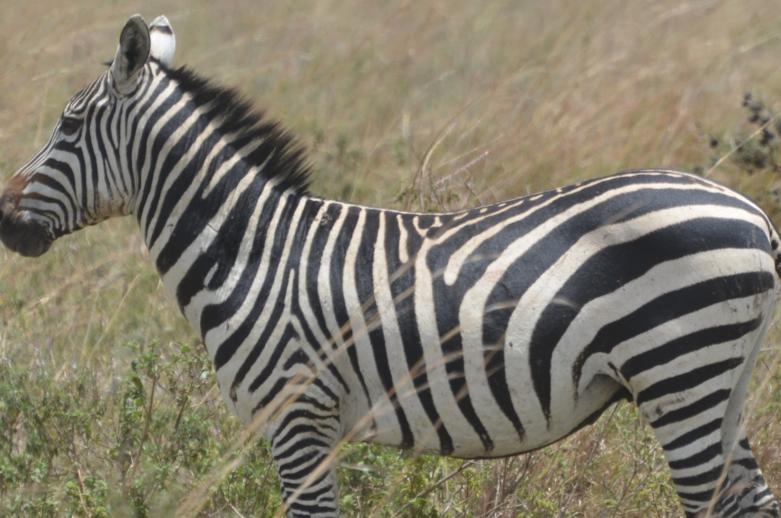} & \includegraphics[height=1.9cm,width=2.3cm]{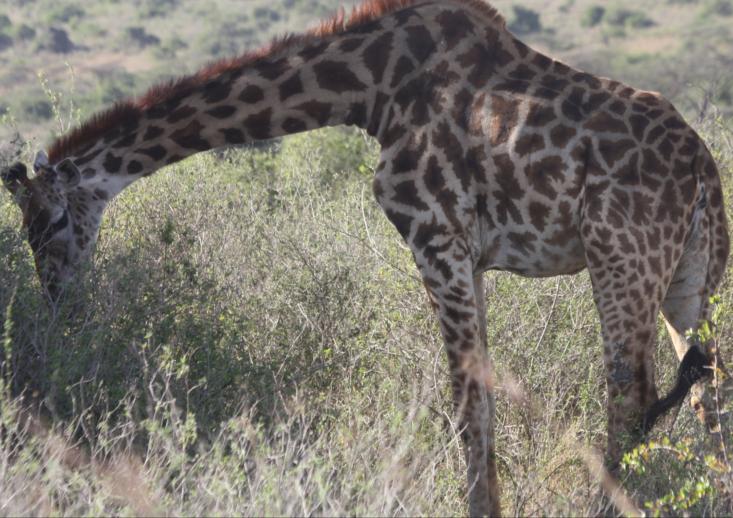} &
    \includegraphics[height=1.9cm,width=2.3cm]{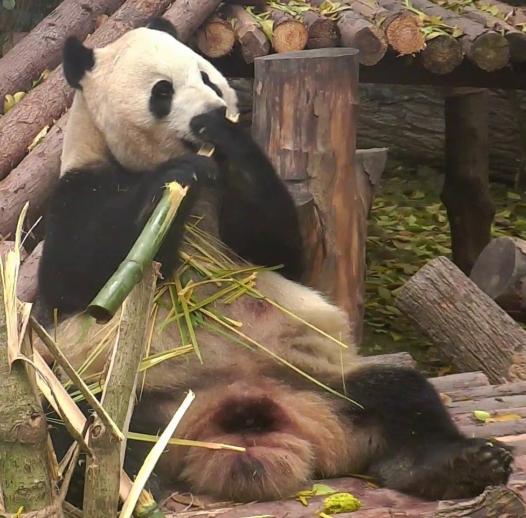} & \includegraphics[height=1.9cm,width=2.3cm]{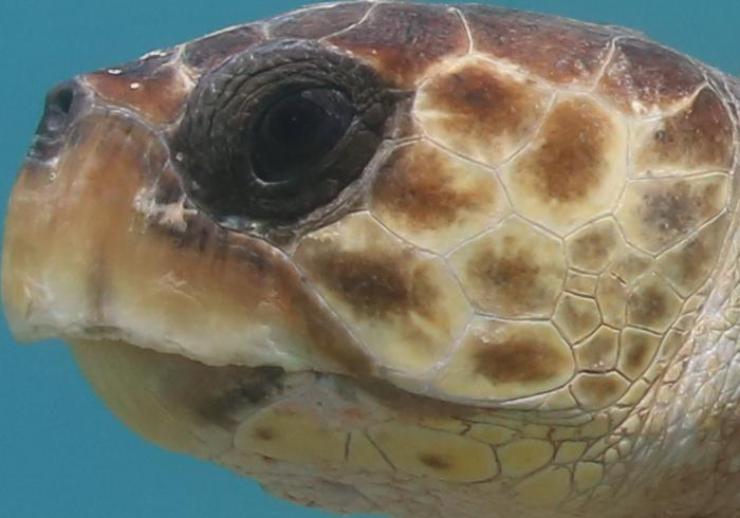}
    \includegraphics[height=1.9cm,width=2.3cm]{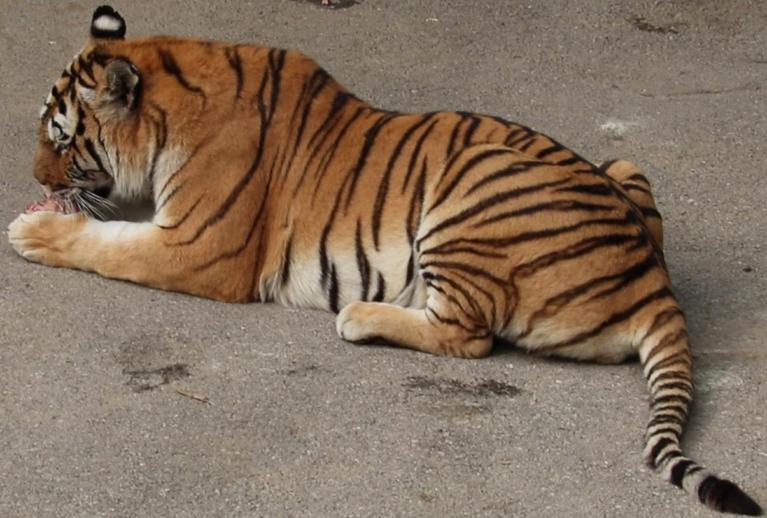}
\end{tabular}
\vspace{-0.1cm}
\caption{
\textbf{Distinct animal features for re-identification}. 
Based on the natural visual appearance, the most distinguishable features for animals are spots, stripes, facial landmarks, and the shape of body parts (e.g., ears for elephants and fin for whales).}
\label{fig:animal_samples}
\vspace{-8pt}
\end{figure}

\section{Datasets}\label{section:datasets}

We evaluate WildFusion on 17 datasets\footnote{For zero-shot, we use only two that were not used in the MegaDescriptor training.} that include diverse species of animals. The datasets were acquired with the help of the recently developed library \href{https://github.com/WildlifeDatasets/wildlife-datasets}{Wildlife Datasets}\cite{Cermak_2024_WACV}, which allows easy access to the datasets and provides unified dataset splits.
We selected subsets of the datasets that are saturated in performance or have a large number of images. Sample images from selected datasets are shown in Figure\,\ref{fig:animal_samples}. For basic statistics of the datasets, see Table~\ref{table:datasets}.

To construct the appropriate training (\textit{database}) and test (\textit{query}) datasets, we followed the methodology proposed in\cite{Cermak_2024_WACV}. However, while analyzing this procedure, we discovered inconsistencies caused by the incorrect loading of images for ATRW and NDD20 datasets due to multiple identities in one image. For these datasets, we fixed the loading by applying the appropriate bounding box or segmentation mask. Therefore, as loaded images are not exactly the same, the achieved results for these two datasets are higher than in the original work.

\begin{table}[!h]
\vspace{-10pt}
\caption{\textbf{Characteristics of selected datasets}. $^\dagger$Used in zero-shot scenario.}
\vspace{-0.2cm}
\small
\setlength{\tabcolsep}{0.65em} 
\centering
\begin{tabular}{lrrr}
\toprule
   & category & \# of images & \# of individuals   \\
\midrule
ATRW\cite{li2019atrw} & tigers  & 5,415 & 182   \\
CowDataset$^\dagger$ \cite{cowdataset} & cows  & 1485 & 13   \\
GiraffeZebraID\cite{parham2017animal} & giraffes, zebras & 6,925 & 2,056   \\
Giraffes\cite{miele2021revisiting} &  giraffes & 1,393 & 178   \\
HyenaID2022\cite{botswana2022} & hyenas & 3,129 & 256  \\
LeopardID2022\cite{botswana2022} & leopards & 6,806 &           430 \\
NyalaData\cite{dlamini2020automated} & nyalas &            1,942 &           237  \\
SealID\cite{nepovinnykh2022sealid} & seals &          2,080 &            57  \\
SeaStarReID2023$^\dagger$ \cite{wahltinez2024open} & starfish & 2187 &  95  \\
SeaTurtleID\cite{adam2024seaturtleid} & sea turtles & 7,774 &  400  \\
WhaleSharkID\cite{holmberg2009estimating} & whale sharks & 7,693 &           543  \\
ZindiTurtleRecall\cite{zinditurtles} & sea turtles &   12,803 &  2,265 \\
BelugaID\cite{belugaid} & belugas & 5,902 & 788 \\
CTai\cite{freytag2016chimpanzee} & chimpanzees & 4,662 & 71 \\
IPanda50\cite{wang2021giant} & pandas & 6,874 & 50  \\
NDD20\cite{trotter2020ndd20} & dolphins & 2,657 & 82  \\
NOAARightWhale\cite{rightwhale} & whales & 4,544 & 447 \\
\bottomrule
\end{tabular}
\label{table:datasets}
\vspace{-15pt}
\end{table}

\section{Experiments}

\noindent\textbf{Evaluation protocol}.
We consider the same closed-set splits as in\cite{Cermak_2024_WACV}, meaning that all individual animals are both in the database (training) and query (test) sets. We approach the problem as image retrieval: for each image in the query set, we find an image in the database and make the query prediction have the same identity as the image from the database. Performance in all experiments is measured as top-1 accuracy.


WildFusion relies on finding hyper-parameters and fitting calibration models. The standard approach involves splitting the training set into development and validation parts, which are used for the selection of the best hyper-parameters and fitting calibration models. However, this approach is not applicable in our case because the MegaDescriptor was trained on the whole training set and cannot be used for validation. We addressed this issue by splitting the original test set into a validation set and a new, smaller test set using a 0.5 ratio. We estimated both $\mu$ and the calibration function on the validation set and utilized them for the final prediction on the test set. Due to this change in test set, our results are not directly comparable to the results reported by\cite{Cermak_2024_WACV}.\vspace{-0.15cm} \\

\noindent\textbf{Technical details}.
To construct the global scores, we use embeddings extracted by MegaDescriptor\cite{Cermak_2024_WACV} and DINOv2\cite{oquab2023dinov2}. For local matching scores, we use LightGlue\cite{lindenberger2023lightglue} feature matching with local descriptors ALIKED\cite{Zhao2023ALIKED}, DISK\cite{tyszkiewicz2020disk}, and SuperPoint\cite{detone2018superpoint}. We use at most 512 keypoints and their appropriate descriptors, extracted from images resized to $512\times512$. For matching with LoFTR\cite{sun2021loftr}, we use the outdoor variant trained on the MegaDepth\cite{li2018megadepth} dataset. On input, we use image pairs with both images resized to $512\times512$. A total of four local feature matching methods were considered to construct matching-based scores. All these methods were taken off-the-shelf, and none were fine-tuned or retrained. We perform the experiments on the datasets described in Section \ref{section:datasets}.
In the baseline WildFusion, we search for optimal hyperparameter $\mu$ from Equation \ref{eq:similarity_local} separately for each dataset. The calibrated scores are given equal weights $w_i$. The summary of settings is in Table \ref{tab:baseline_method}. \vspace{-0.15cm}

\begin{table}[h!]
\centering
\caption{\textbf{WilfFusion settings overview}. We test a variety of state-of-the-art local and global methods for animal re-identification and image retrieval. The calibration is done using Logistic or Isotonic regression.}\vspace{-0.15cm}
\begin{tabular}{l|@{\hspace{0.4cm}}l@{\hspace{0.4cm}}l}
\toprule
\textbf{Components:  } & Local matching methods: & Global similarity methods:   \\
    & ~-- LoFTR & ~-- MegaDescriptor-L-384 \\
    & ~-- LightGlue + SuperPoint & ~-- DINOv2-512\\
    & ~-- LightGlue + Disk & \\
    & ~-- LightGlue + Aliked & \\
\midrule
\textbf{Calibration:} & \multicolumn{2}{l}{Isotonic regression with PCHIP interpolation} \\
 & \multicolumn{2}{l}{Logistic regression} \\
\midrule
\textbf{Fusion:} & Average with equal weights $w_i$ & \\

\bottomrule
\end{tabular}
\label{tab:baseline_method}
\end{table}

\subsection{Baseline Performance}

WildFusion clearly outperforms MegaDescriptor in most scenarios. When using all available scores, WildFusion shows superior performance on 16 out of 17 datasets, with only one dataset, ZindiTurtleRecall, showing a slight decrease in accuracy. The average accuracy improvement is substantial, with WildFusion (all) achieving 84.0\% compared to MegaDescriptor's 75.5\%, representing a notable average gain of 8.5 percentage points. The most significant improvements are seen in datasets like NDD20, WhaleSharkID, SeaStarReID2023, and SealID, where WildFusion shows accuracy gains of over 14 percentage points.

Interestingly, even when using only local matching scores, WildFusion maintains competitive performance. It outperforms MegaDescriptor on 11 out of 17 datasets with average accuracy (78.5\%), which is better than MegaDescriptor by 3.0 percentage points. This suggests that the local matching scores are quite powerful on their own, without any need for fine-tuning on animal datasets. More details about the results, including per dataset performance, are in Table~\ref{tab:main}. Besides, we provide a qualitative evaluation in Figure \ref{fig:nyala_data_grid}

\input{tables/table-main-results}

\input{tables/figure}

\newpage
\section{Ablation Studies}

This section presents a set of ablation studies to empirically validate the design choices behind the WildFusion.

\subsection{Effect of local matching score threshold}
\label{sec:effect-mu}
\begin{wrapfigure}{r}[0pt]{0.45\textwidth}
  \vspace{-0.85cm}
  \includegraphics[width=0.45\textwidth]{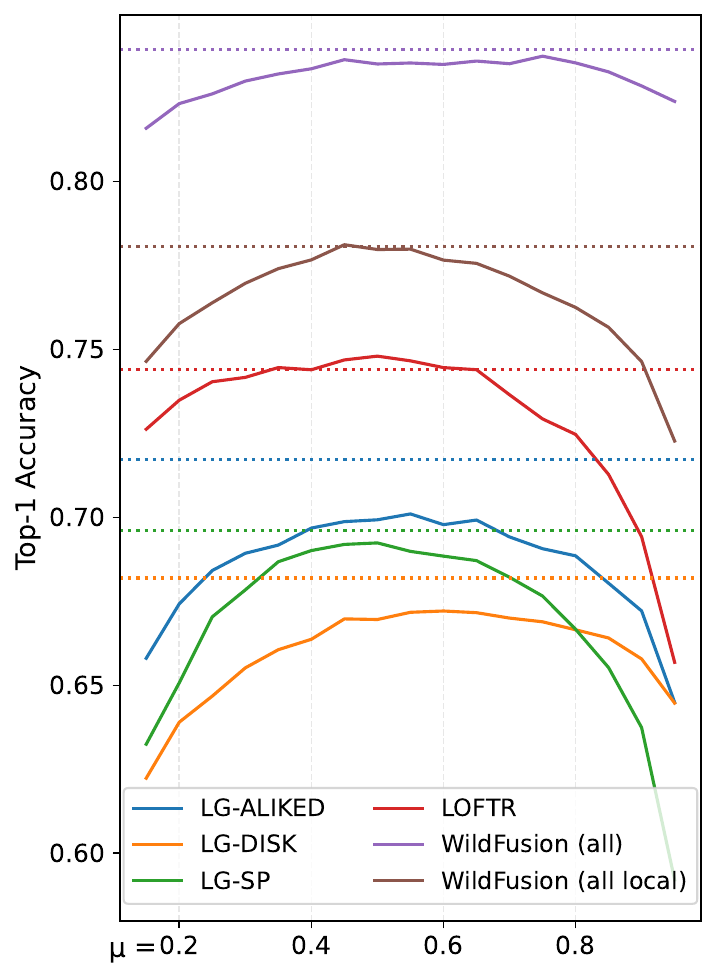} 
  \vspace{-0.5cm}
  \caption{\textbf{Effect of $\mu$ on performance.} Full lines represents constant $\mu$, and dotted lines optimal $\mu$ found on validations set for each dataset. Fixing $\mu=0.5$ provides comparable results to the best $\mu$ based on validation set.}
  \label{fig:effect-mu}
  \vspace{-0.5cm}
\end{wrapfigure}

The hyperparameter $\mu$ controls the trade-off between low-quality matches and fewer high-quality matches. When $\mu$ is low, the score is influenced by many low-quality matches, often presented in the background. When $\mu$ is very high, it filters out most of the matches, leading to a loss of information and resulting in a zero score for nearly all pairs.

Comparing performance scores with constant $\mu$  and $\mu$ selected based on the validation set suggests that local methods are robust to $\mu$ selection, and selecting any $\mu$ values between [0.4, 0.6] is a good choice. Interestingly, local methods perform better with $\mu$=0.45 fixed for all datasets than searching for optimal $\mu$ on the validation set. When local matching scores are combined with global scores, the range of suitable $\mu$ values is wider and extends from 0.4 to 0.8. This shows that adding global scores to the ensemble reduces the downside of having zeros in the score for large $\mu$ values (see Figure \ref{fig:effect-mu}). 

\subsection{Effect of score selection}
WildFusion's versatility allows it to fuse any score. As mentioned, using WildFusion only with all local matching scores outperforms the \mbox{MegaDescriptor-L} global score. When we included the global score from the general-purpose feature extractor DinoV2, performance improved only marginally, highlighting the importance of fine-tuning the deep embedding model.

Using the MegaDescriptor-L global score with at least one local matching score significantly outperforms using MegaDescriptor-L alone. Combining it with LG-ALIKED achieves the highest accuracy of 83.0\%, followed by LoFTR at 81.4\%. LG-SuperPoint and LG-DISK also show comparable performance with accuracies of 80.6\% and 81.1\%, respectively. Combining the global score with all local matching scores further improves performance, suggesting that the local matching scores are mostly uncorrelated and perform well in the ensemble. More details can be found in Table \ref{tab:wildfusion-variety}.

\input{tables/table-variants}

\subsection{Effect of calibration}
\vspace{-0.2cm}
Using isotonic regression for calibration yields marginally better results on average compared to logistic regression (83.9\% accuracy). However, there is a discrepancy in performance between the datasets. For example, using logistic regression was better on NDD20 (+3.0\%) and ZindiTurtleRecall (+ 2.7\%), but it significantly underperformed on NyalaData (-6.5\%) compared to the isotonic regression. This suggests that the poor performance of WildFusion on ZindiTurtleRecall can be related to incorrect calibration. \vspace{-0.1cm}\\

\begin{figure}[h]
\vspace{-0.5cm}
\centering
\includegraphics[scale=0.5]{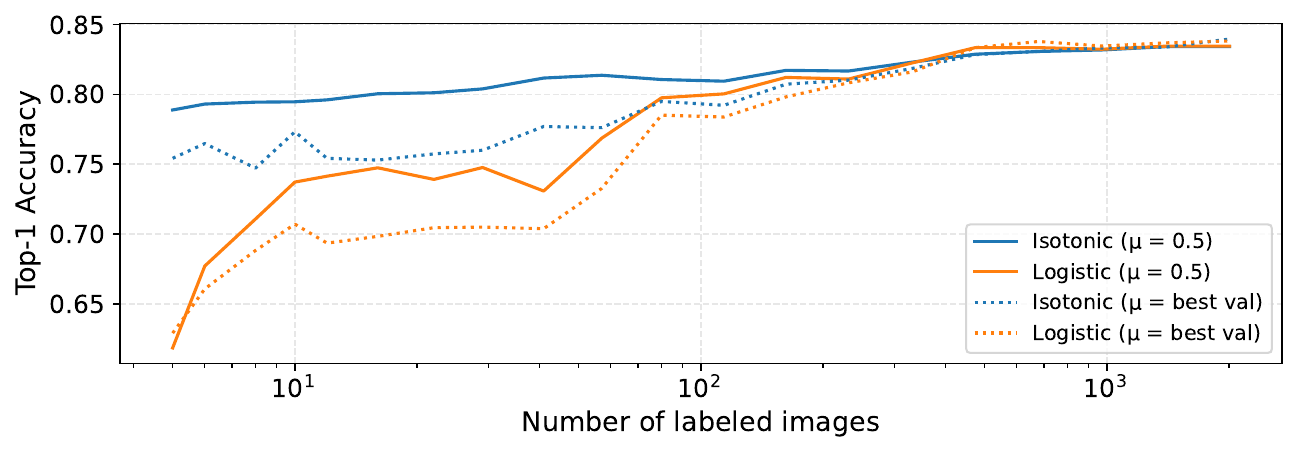}
\vspace{-0.2cm}
\caption{
\textbf{Ablation on optimal number of images for model calibration.}
Isotonic regression calibration with fixed $\mu=0.5$ for all datasets outperforms other approaches in low data scenarios. }
\label{fig:data-calibration}
\vspace{-0.3cm}
\end{figure}

\noindent\textbf{How much data do we need for calibration?}
To test this, we create variously sized subsets of labeled images from database and validation sets, such that at least 2 positive and 2 negative pairs can be created. Pairs created from this subset are used for calibration and finding $\mu$. We perform additional experiments with $\mu$ fixed to 0.5 to isolate the effect of low data calibration from finding~$\mu$.
As visualized in Figure \ref{fig:data-calibration}, isotonic regression performs better than logistic regression in low data scenarios, both with optimized and fixes $\mu$. Calibration with fixed $\mu$ is significantly better for a smaller number of samples but yields marginally worse results when a lot of labeled data is available. In general, calibration is very data efficient. For example, fixing $\mu$ to 0.5 and calibrating each dataset using only 10 labeled images still gives a reasonable 79.5\% accuracy. Adding more data to the calibration further increases the performance of up to 200 samples, where additional data only gives marginal improvements. Our results suggest that WildFusion is viable even with very few labeled samples.

\subsection{Constraining number of comparisons}
Given a database with $M$ samples and a query with $N$ samples, methods based on local features often need to perform pairwise comparisons, needing $M\times N$ comparisons. 
Since many modern matching algorithms are based on neural networks $ M\times N$, neural network inferences are required. 
With the increasing database size, the computational time quickly becomes unfeasible; therefore, the calculation of all local scores remains a viable option only for moderately sized datasets. \\

\noindent\textbf{Shortlist strategy}.
We consider a scenario where we have two types of scores, one cheap to calculate, such as global score  $s_{G}$  from MegaDescriptor or DinoV2, and one expensive, such as WildFusion with local matching scores $s_{W}$. We follow the shortlist strategy\cite{yao2017large} and use cheap global scores to filter candidate samples. The expensive scores are calculated for a restricted size shortlist  to validate and re-rank the top matches. The running time is controlled by a computational budget $B$ in terms of the number of expensive score evaluations per query image. \\

\begin{figure}[!h]
\centering
\includegraphics[scale=0.5]{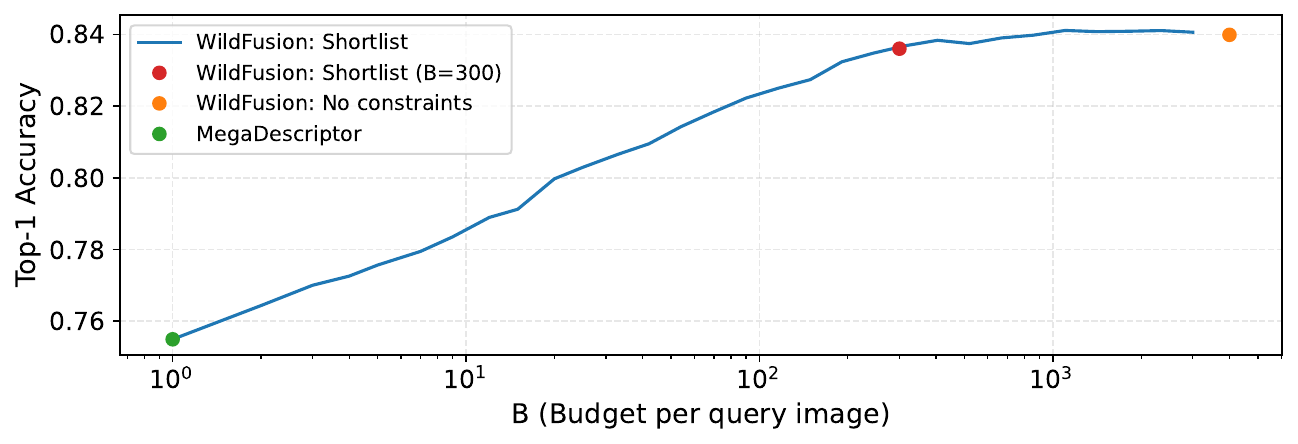}
\caption{\textbf{Rate of performance improvement with increasing budget}. The shortlist strategy allows adding more computational resources to improve performance, up to a budget of $B=200$.}
\label{fig:priority}
\end{figure}

\noindent\textbf{Results}. Using the shortlist strategy, we are efficiently able to utilize the WildFusion scores $s_{W}$, which are costly to calculate. On average, budget $B=300$ is enough to reach accuracy comparable to calculating all scores. For example, on SeaTurtleIdHeads, WildFusion needs only about 200 $s_{W}$ calculations to reach its peak performance. With a database size of 6063, this results in more than a 30-fold increase in inference speed. Interestingly, performance at $B=10^{3}$ is slightly better than using all comparisons. It indicates that local matching scores in WildFusion are prone to some degree of false positive matches when applied to all images in the database. A more detailed visualization of the speed-up is in Figure \ref{fig:priority}.

\section{Zero shot performance}
Encouraged by the fact that calibration works well even with a low number of data points to achieve reasonable performance, we conducted an experiment in a zero-shot setting, meaning no data is needed prior to inference. We split datasets into disjoint subsets. For each score, we trained a single calibration model on one subset and evaluated it on a different subset, with $\mu=0.5$ fixed for all local matching scores. This differs from the default setting, where subsets of the same dataset were used for calibration and evaluation.

For the zero-shot experiment, we evaluated WildFusion using only local matching scores without incorporating MegaDescriptor's global scores, as the latter had already been trained on the data. Our zero-shot WildFusion approach achieved an average accuracy of 76.2\%, which is 2.3 percentage points lower than the accuracy obtained with dataset-specific calibration. Notably, this performance is also 0.7 percentage points higher than the state-of-the-art fine-tuned model MegaDescriptor-L-384, demonstrating the effectiveness of our method in a zero-shot setting without fine-tuning or dataset-specific calibration. For more detail about performance, see Table \ref{tab:wildfusion-zero-shot}.

For both novel datasets not included in the MegaDescriptor training set, WildFusion with local scores achieved a perfect 100\% accuracy. In contrast, for the CowsDataset, DINOv2 reached an accuracy of 96.0\%, while MegaDescriptor achieved 98.7\%. Similarly, for the SeaStarReID2023 dataset, DINOv2 obtained an accuracy of 82.2\%, whereas MegaDescriptor reached 88.8\%.

\begin{table}[h!]
\setlength{\tabcolsep}{4.85pt} 
\renewcommand{\arraystretch}{1.25}

\centering
\vspace{-0.25cm}
\caption{\textbf{WildFusion performance in zero-shot setting}. No data from the evaluated dataset was used prior to test time (except Wahltinez et al.\cite{wahltinez2024open}, which used standard classification setting).}
\vspace{-0.1cm}
\begin{tabular}{lcccc}
\toprule
&  &  & (\textit{local})  &\cite{wahltinez2024open}\\
& MegaDescriptor-L & DINOv2 & WildFusion  & Wahltinez et al. \\
\midrule
CowsDataset     & \colorbox{LightGreen}{98.7}  & 88.8  & \colorbox{Greener}{100.0} & -- \\
SeaStarReID2023 & 82.2 & 96.0  & \colorbox{Greener}{100.0} & \colorbox{LightGreen}{99.9} \\
\midrule
17 datasets   & -- & 47.5 & 76.2 & --  \\

\bottomrule
\end{tabular}
\label{tab:wildfusion-zero-shot}
\vspace{-0.7cm}
\end{table}

\section{Conclusion}
In this paper, we presented WildFusion, a novel approach to individual animal identification that leverages a calibrated similarity fusion of deep and local matching scores. By combining deep features extracted from MegaDescriptor or DINOv2 with local matching descriptors (e.g., LoFTR and SuperPoint), WildFusion achieves state-of-the-art performance across a wide range of datasets.
Our method is easy to use in real applications as it does not require training and is usable out of the box with any pre-trained deep embedding models and local feature-matching methods. Besides, the code was made public.

Even though the best results were obtained with dataset-specific calibration, we have empirically shown that using WildFusion of only local similarity score and with generic calibration still gives good performance, with mean accuracy dropping only by 2.3\% and still reduced the relative error of MegaDescriptor by 44 percentage points.
WildFusion's flexibility was also further proven by its strong performance in zero-shot settings tested on species "never seen before." 

The scalability and generalization potential of WildFusion makes it suitable for application across different species and environments, contributing significantly to the field of animal re-identification. \\ 

\noindent\textbf{Limitations:}
WildFusion leverages off-the-shelf local matching methods like LoFTR and LightGlue, which were originally trained on datasets featuring static objects. Therefore it is not optimized for matching animals, where the same animal can be observed in various poses,  lighting conditions and with occlusions. Therefore, our work can be extended by adapting or retraining these local feature-matching models specifically for animal identification tasks, potentially improving the accuracy and robustness of the WildFusion approach. The same applies to deep descriptors such as MegaDescriptor and DINOv2, which, if not trained on that species, will most likely underperform.

WildFusion is best suited for offline analysis of existing databases. While we introduced a method to address scalability, it remains insufficient for real-time identification. Future research could explore the development of more efficient algorithms to enable real-time processing and online identification.

\section*{Acknowledgements}
The authors were supported by the Technology Agency of the Czech Republic, project No. SS05010008.
Computational resources were provided by the e-INFRA CZ project (ID:90254), supported by the Ministry of Education, Youth and Sports of the Czech Republic.

%
%
\bibliographystyle{splncs04}
\bibliography{main}

\end{document}

%% file: tables/table-main-results.tex

\begin{table}[h]
  \setlength{\tabcolsep}{0.85em}
\centering
\caption{\textbf{WildFusion's performance in comparison with MegaDescriptor}. On average, WildFusion, outperforms MegaDescriptor, even with just \textit{local} descriptors. WildFusion with \textit{all} local and deep descriptors ranks the best on all but two datasets.}
\label{tab:main}
\begin{tabular}{lcc@{\hspace{-5pt}}rc@{\hspace{-5pt}}r}
\toprule
 & MegaDescriptor & (\textit{all}) &  & (\textit{local}) &  \\
 & Large-384 & WildFusion & $\Delta$~~ & WildFusion & $\Delta$~~ \\
\midrule
ZindiTurtleRecall & \colorbox{Greener}{74.24} & \colorbox{LightGreen}{71.90} & \text{\small -2.34} & 45.62 & \text{\small -28.62} \\
CTai & \colorbox{LightGreen}{91.86} & \colorbox{Greener}{92.08} & \text{\small +0.21} & 81.80 & \text{\small -10.06} \\
ATRW & 97.96 & \colorbox{Greener}{98.51} & \text{\small +0.56} & \colorbox{LightGreen}{98.33} & \text{\small +0.37} \\
CowDataset & \colorbox{LightGreen}{98.66} & \colorbox{Greener}{100.00}~\, & \text{\small +1.34} & \colorbox{Greener}{100.00}~\, & \text{\small +1.34} \\
SeaTurtleIDHeads & 91.18 & \colorbox{Greener}{95.00} & \text{\small +3.82} & \colorbox{LightGreen}{93.82} & \text{\small +2.63} \\
IPanda50 & \colorbox{LightGreen}{85.76} & \colorbox{Greener}{89.68} & \text{\small +3.92} & 81.40 & \text{\small -4.36} \\
NyalaData & \colorbox{LightGreen}{41.59} & \colorbox{Greener}{46.26} & \text{\small +4.67} & 25.23 & \text{\small -16.36} \\
BelugaID & \colorbox{LightGreen}{67.61} & \colorbox{Greener}{72.46} & \text{\small +4.85} & 63.07 & \text{\small -4.54} \\
NOAARightWhale & \colorbox{LightGreen}{43.25} & \colorbox{Greener}{49.25} & \text{\small +6.00} & 42.18 & \text{\small -1.07} \\
Giraffes & 91.04 & \colorbox{Greener}{99.25} & \text{\small +8.21} & \colorbox{LightGreen}{98.51} & \text{\small +7.46} \\
HyenaID2022 & 78.41 & \colorbox{Greener}{90.48} & \text{\small +12.06} & \colorbox{LightGreen}{88.25} & \text{\small +9.84} \\
GiraffeZebraID & 82.98 & \colorbox{Greener}{95.74} & \text{\small +12.77} & \colorbox{LightGreen}{94.81} & \text{\small +11.84} \\
LeopardID2022 & 77.82 & \colorbox{Greener}{90.93} & \text{\small +13.11} & \colorbox{LightGreen}{89.40} & \text{\small +11.58} \\
SealID & 78.47 & \colorbox{Greener}{92.82} & \text{\small +14.35} & \colorbox{LightGreen}{90.91} & \text{\small +12.44} \\
SeaStarReID2023 & 82.24 & \colorbox{LightGreen}{99.53} & \text{\small +17.29} & \colorbox{Greener}{100.00} & \text{\small +17.76} \\
WhaleSharkID & 62.04 & \colorbox{Greener}{80.33} & \text{\small +18.28} & \colorbox{LightGreen}{77.68} & \text{\small +15.64} \\
NDD20 & 38.35 & \colorbox{Greener}{63.53} & \text{\small +25.19} & \colorbox{LightGreen}{63.16} & \text{\small +24.81} \\
\midrule
\textit{Average} & \textit{75.50} & \colorbox{Greener}{\textit{83.99}} & \text{\small \textit{+8.49}} & \colorbox{LightGreen}{\textit{78.48}} & \text{\small \textit{+2.98}} \\
\bottomrule
\end{tabular}
\end{table}


%% file: tables/figure.tex



\begin{figure}[htbp]
    \centering
    \label{fig:nyala_data_grid}
    \begin{tabular}{@{}c@{\hspace{2pt}}c c c c c c }
        \rotatebox{90}{\scriptsize{~~~~~Query}} & 
        \includegraphics[width=0.15\textwidth]{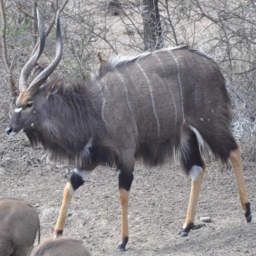} & 
        \includegraphics[width=0.15\textwidth]{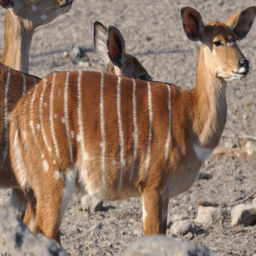} & 
        \includegraphics[width=0.15\textwidth]{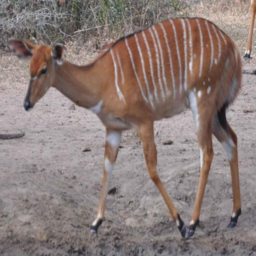} & 
        \includegraphics[width=0.15\textwidth]{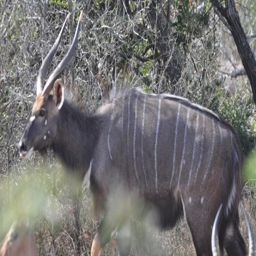} & 
        \includegraphics[width=0.15\textwidth]{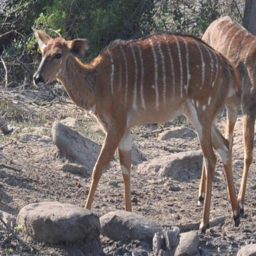} &
        \includegraphics[width=0.15\textwidth]{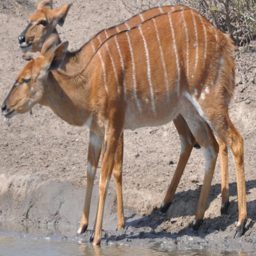} \\
        
        \rotatebox{90}{\scriptsize{~WildFusion}} & 
        \greenbox{\includegraphics[width=0.14\textwidth]{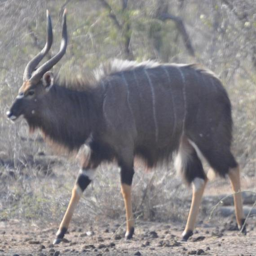}} & 
        \greenbox{\includegraphics[width=0.14\textwidth]{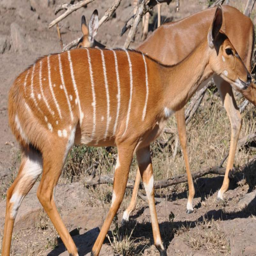}} & 
        \greenbox{\includegraphics[width=0.14\textwidth]{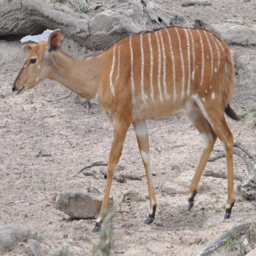}} & 
        \redbox{\includegraphics[width=0.14\textwidth]{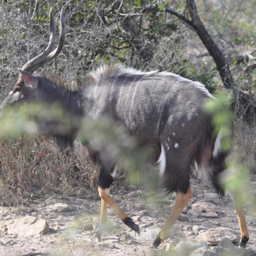}} & 
        \redbox{\includegraphics[width=0.14\textwidth]{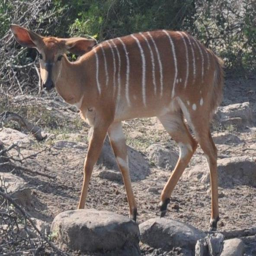}} &
        \redbox{\includegraphics[width=0.14\textwidth]{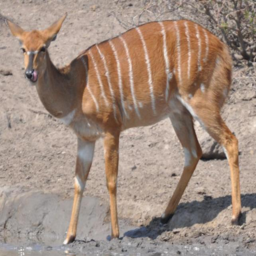}} \vspace{2pt}\\

        \rotatebox{90}{\scriptsize{~MegaDes-L}} & 
        \redbox{\includegraphics[width=0.14\textwidth]{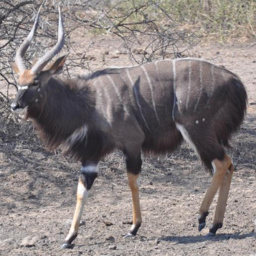}} & 
        \redbox{\includegraphics[width=0.14\textwidth]{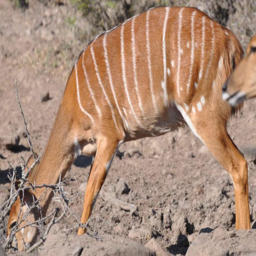}} & 
        \redbox{\includegraphics[width=0.14\textwidth]{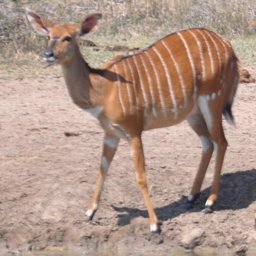}} & 
        \greenbox{\includegraphics[width=0.14\textwidth]{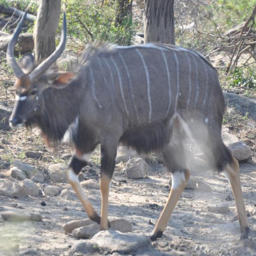}} & 
        \greenbox{\includegraphics[width=0.14\textwidth]{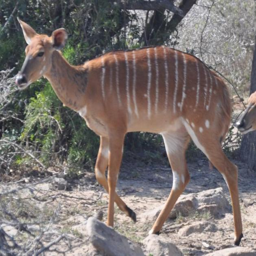}} &
        \greenbox{\includegraphics[width=0.14\textwidth]{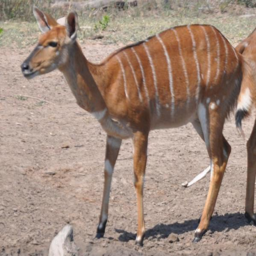}} \vspace{8pt} \\
        
        \rotatebox{90}{\scriptsize{~~~~~~Query}} & 
        \includegraphics[width=0.15\textwidth]{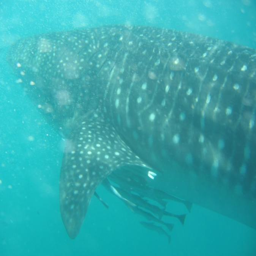} & 
        \includegraphics[width=0.15\textwidth]{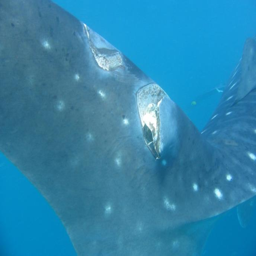} & 
        \includegraphics[width=0.15\textwidth]{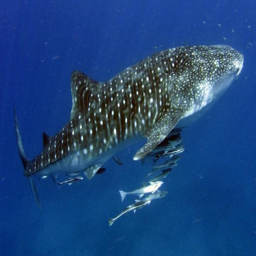} & 
        \includegraphics[width=0.15\textwidth]{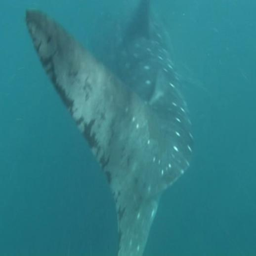} & 
        \includegraphics[width=0.15\textwidth]{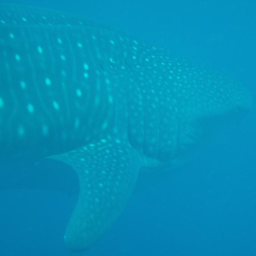} &
        \includegraphics[width=0.15\textwidth]{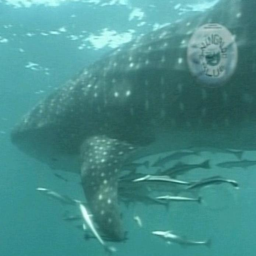} \\

        \rotatebox{90}{\scriptsize{~WildFusion}} & 
        \greenbox{\includegraphics[width=0.14\textwidth]{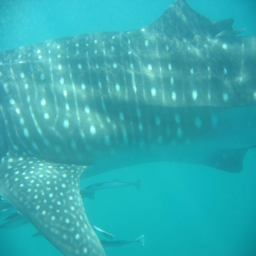}} & 
        \greenbox{\includegraphics[width=0.14\textwidth]{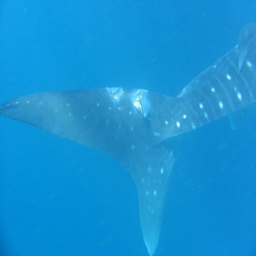}} & 
        \greenbox{\includegraphics[width=0.14\textwidth]{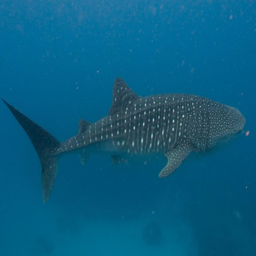}} & 
        \redbox{\includegraphics[width=0.14\textwidth]{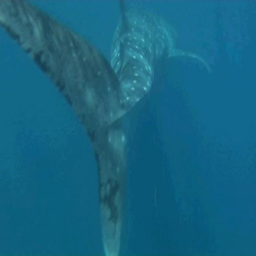}} & 
        \redbox{\includegraphics[width=0.14\textwidth]{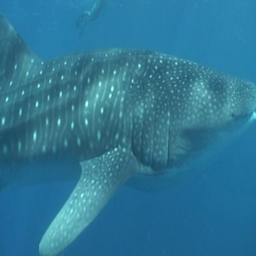}} &
        \redbox{\includegraphics[width=0.14\textwidth]{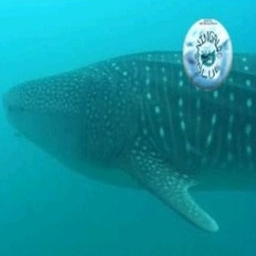}} \vspace{2pt}\\

        \rotatebox{90}{\scriptsize{~~MegaDes-L}} & 
        \redbox{\includegraphics[width=0.14\textwidth]{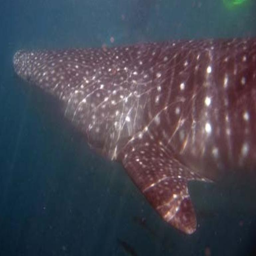}} & 
        \redbox{\includegraphics[width=0.14\textwidth]{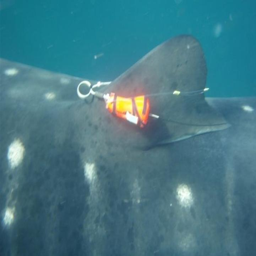}} & 
        \redbox{\includegraphics[width=0.14\textwidth]{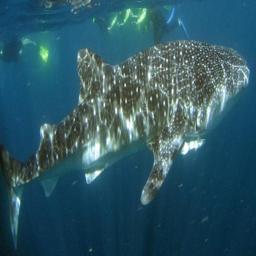}} & 
        \greenbox{\includegraphics[width=0.14\textwidth]{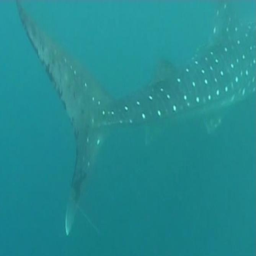}} & 
        \greenbox{\includegraphics[width=0.14\textwidth]{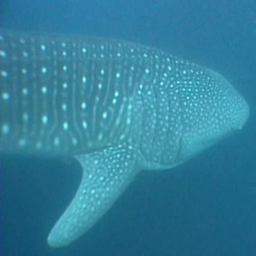}} &
        \greenbox{\includegraphics[width=0.14\textwidth]{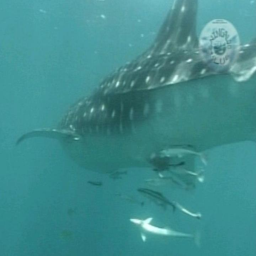}} \vspace{8pt} \\

        \rotatebox{90}{\scriptsize{~~~~~~Query}} & 
        \includegraphics[width=0.15\textwidth]{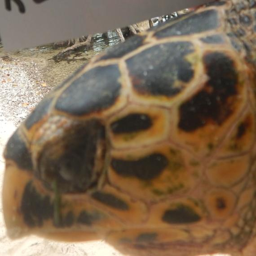} & 
        \includegraphics[width=0.15\textwidth]{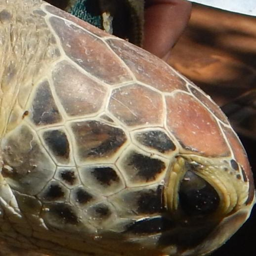} & 
        \includegraphics[width=0.15\textwidth]{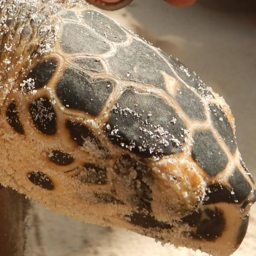} & 
        \includegraphics[width=0.15\textwidth]{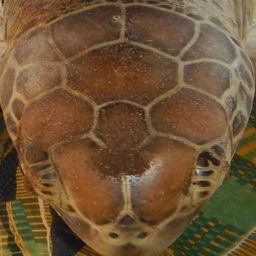} & 
        \includegraphics[width=0.15\textwidth]{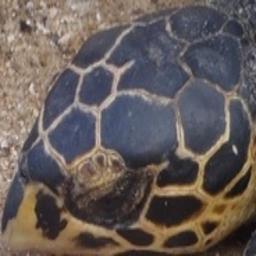} &
        \includegraphics[width=0.15\textwidth]{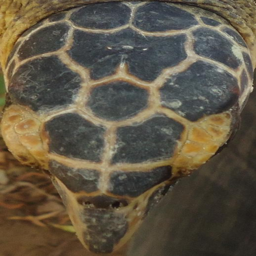} \\

        \rotatebox{90}{\scriptsize{~WildFusion}} & 
        \greenbox{\includegraphics[width=0.14\textwidth]{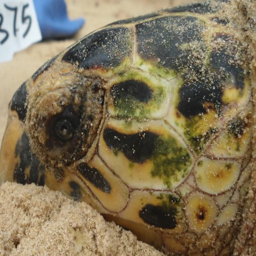}} & 
        \greenbox{\includegraphics[width=0.14\textwidth]{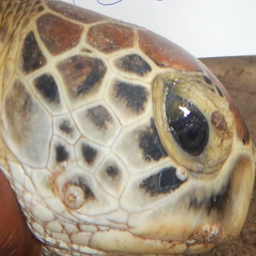}} & 
        \greenbox{\includegraphics[width=0.14\textwidth]{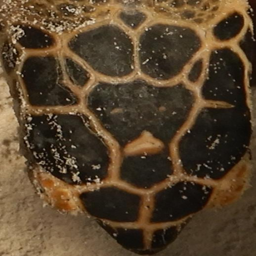}} & 
        \redbox{\includegraphics[width=0.14\textwidth]{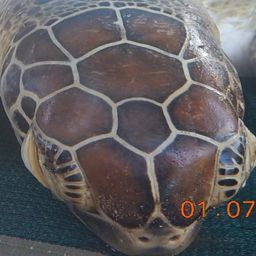}} & 
        \redbox{\includegraphics[width=0.14\textwidth]{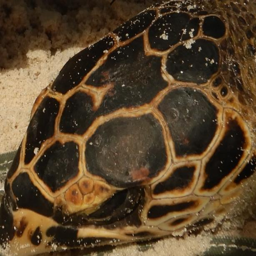}} &
        \redbox{\includegraphics[width=0.14\textwidth]{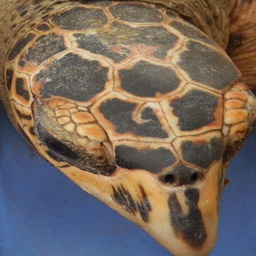}} \vspace{2pt}\\

        \rotatebox{90}{\scriptsize{~MegaDes-L}} & 
        \redbox{\includegraphics[width=0.14\textwidth]{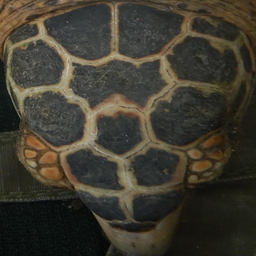}} & 
        \redbox{\includegraphics[width=0.14\textwidth]{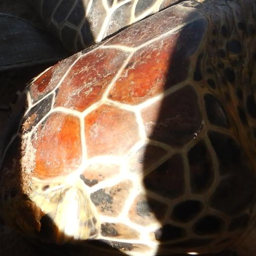}} & 
        \redbox{\includegraphics[width=0.14\textwidth]{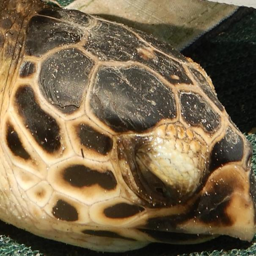}} & 
        \greenbox{\includegraphics[width=0.14\textwidth]{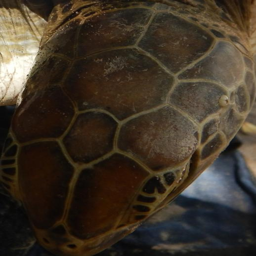}} & 
        \greenbox{\includegraphics[width=0.14\textwidth]{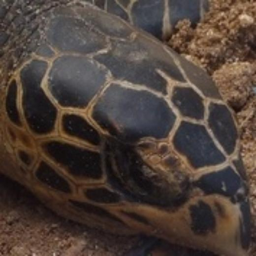}} &
        \greenbox{\includegraphics[width=0.14\textwidth]{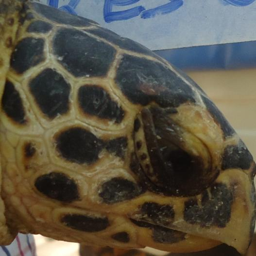}} \\
    \end{tabular}
        \caption{\textbf{Qualitative performance.} Selected examples where WildFusion changed the decision of the MegaDescriptor-L on NyalaData, WhaleSharkID, and ZindiTurtle; three correct and false samples. We suspect that some wrong matches are mislabeled data.}
    \label{fig:nyala_data_grid}

\end{figure}




%% file: tables/table-variants.tex

\begin{table}[h!]
\setlength{\tabcolsep}{4.85pt} 
\renewcommand{\arraystretch}{1.25}

\centering
\caption{\textbf{Ablation on local and global score fusion}. We report WildFusion's performance using various local and global methods. Combining local methods with fine-tuned global scores of MegaDetector-L achieves the best results.}
\vspace{-0.2cm}
\begin{tabular}{lcccccc}
\toprule
\diagbox{Global}{Local} & \textit{None} & LG-DISK & LG-SP & LG-ALIKED & LoFTR & all \\
\midrule
\textit{None} & - & 68.9 & 70.1 & 72.2 & 74.8 & 78.5 \\
DINOv2 & 47.5 & 70.4 & 71.6 & 73.7 & 74.8 & 78.8 \\
MegaDescriptor-L & 75.5 & 81.1 & 80.6 & \colorbox{LightGreen}{83.0} & 81.4 & \colorbox{Greener}{84.0} \\
\bottomrule
\end{tabular}
\label{tab:wildfusion-variety}
\vspace{-0.4cm}
\end{table}
